\documentclass[conference]{IEEEtran}
\IEEEoverridecommandlockouts
\ifCLASSINFOpdf
\else
\fi
\usepackage{multirow}
\usepackage{comment}
\usepackage{array}
\usepackage[acronym,toc,shortcuts]{glossaries}
\usepackage[usenames, dvipsnames]{color}
\usepackage{url}
\usepackage{csquotes}
\usepackage{amssymb,amsmath}
\usepackage{relsize}
\DeclareMathOperator*{\argmax}{arg\,max}

\usepackage{float}
\usepackage[acronym,toc,shortcuts]{glossaries}
\usepackage{graphicx}
\usepackage{caption}
\usepackage{subcaption}
\usepackage{soul}
\usepackage{algorithm,algcompatible}
\usepackage[noend]{algpseudocode}
\usepackage{enumitem}
\usepackage{booktabs,tabularx}
\usepackage{bbm}

\usepackage{upgreek}

\IEEEoverridecommandlockouts
\newcommand{\matr}[1]{\mathbf{#1}}
\begin{document}

%
\title{Transfer Learning of an Ensemble of DNNs for SSVEP BCI Spellers without User-Specific Training
}



%
\author{\IEEEauthorblockN{Osman Berke Guney and
Huseyin Ozkan,\IEEEmembership{~Member,~IEEE} }
\thanks{O. B. Guney (corresponding author) is with the Department of Electrical and Computer Engineering at Boston University, Boston, MA, USA (e-mail: berke@bu.edu). H. Ozkan is with the Faculty of Engineering and Natural Sciences at Sabanci University, Istanbul, Turkey (e-mail: huseyin.ozkan@sabanciuniv.edu).}}


\maketitle
\thispagestyle{plain}
\pagestyle{plain}


\begin{abstract}

\textit{Objective:} Steady-state visually evoked potentials (SSVEPs), measured with EEG (electroencephalogram), yield decent information transfer rates (ITR) in brain-computer interface (BCI) spellers. However, the current high performing SSVEP BCI spellers in the literature require an initial lengthy and tiring user-specific training for each new user for system adaptation, including data collection with EEG experiments, algorithm training and calibration (all are before the actual use of the system). This impedes the widespread use of BCIs. To ensure practicality, we propose a highly novel target identification method based on an ensemble of deep neural networks (DNNs), which does not require any sort of user-specific training. \textit{Method:} We exploit already-existing literature datasets from participants of previously conducted EEG experiments to train a global target identifier DNN first, which is then fine-tuned to each participant. We transfer this ensemble of fine-tuned DNNs to the new user instance, determine the $k$ most representative DNNs according to the participants' statistical similarities to the new user, and predict the target character through a weighted combination of the ensemble predictions.  \textit{Results:} On two large-scale benchmark and BETA datasets, our method achieves impressive \textit{155.51 bits/min} and \textit{114.64 bits/min} ITRs. Code is available for reproducibility: {https://github.com/osmanberke/Ensemble-of-DNNs} \textit{Conclusion:} The proposed method significantly outperforms all the state-of-the-art alternatives for all stimulation durations in $[0.2-1.0]$ seconds on both datasets. \textit{Significance:} Our Ensemble-DNN method has the potential to promote the practical widespread deployment of BCI spellers in daily lives as we provide the highest performance while enabling the immediate system use without any user-specific training.
\end{abstract}


\begin{IEEEkeywords}
Brain computer interfaces, BCI, EEG, SSVEP, Ensemble, Deep learning, Transfer learning
\end{IEEEkeywords}

\maketitle
\IEEEpeerreviewmaketitle
\IEEEpubidadjcol
\def\figureautorefname{Fig.}
\section{Introduction} \label{sec:IN}
Brain-computer interfaces (BCIs) enable motor disabled individuals to communicate with and control their surroundings \cite{9023382, zhang2017bci}. BCIs can also be used for stroke rehabilitation \cite{9055030}. Additionally, novel stimulation paradigms \cite{8954814} as well as advanced signal processing and machine learning techniques \cite{kalaganis2019riemannian} have been extensively studied to improve BCIs. A prominent research direction focuses on speller designs in which the user can type alphanumeric characters through brain signals alone by using a visual interface \cite{fbcca}. Electroencephalogram (EEG) is widely used in BCI applications to measure the brain activity non-invasively, providing a direct transmission pathway between a computer and the brain \cite{6807693}. On the other hand, thanks to its high signal-to-noise ratio (SNR), reduced training time and high information transfer rate (ITR), SSVEP (steady state visually evoked potentials) has attracted attention in the field recently and is now one of the leading paradigms \cite{fbcca, 6807693}. SSVEP, measured by EEG, is the brain's response to a visual stimulus calibrated at a constant frequency. The spectrum of the resulting EEG waveform includes strong components that are tuned to the frequency of the input stimulus and its harmonics \cite{ourdnn}.

In SSVEP BCI spellers, the target character identification relies on frequency tagging that refers to the assignment of distinct oscillation frequencies to the visuals of multiple alphanumeric characters. A person sees a matrix of flickering alphanumeric characters (i.e., uniquely tagged with frequencies) on the computer screen. S/he is asked to select and focus on the target character that is to be spelled at that moment. The objective is to correctly predict the target character based on the received SSVEP EEG signal. 

One of the main concerns of the SSVEP BCI speller studies is to ease the system design for convenient implementations in real life. It is naturally not practical to collect data from each new-coming user (with lengthy and tiring preparation or calibration procedures or separate EEG experiments, all for algorithm training and system adaptation) before s/he actually starts using the system. We call these additional processes altogether \textbf{user-specific training} throughout the paper. Therefore, methods that do not require user-specific training are certainly preferable \cite{train-or-not}. Nevertheless, one typically has some existing SSVEP EEG speller data prior to a new user arrives. This data might be from previous users, or persons participated in previous experiments or from the literature benchmarks.
\begin{figure*}[t!]
\centering
\includegraphics[width=0.85\textwidth]{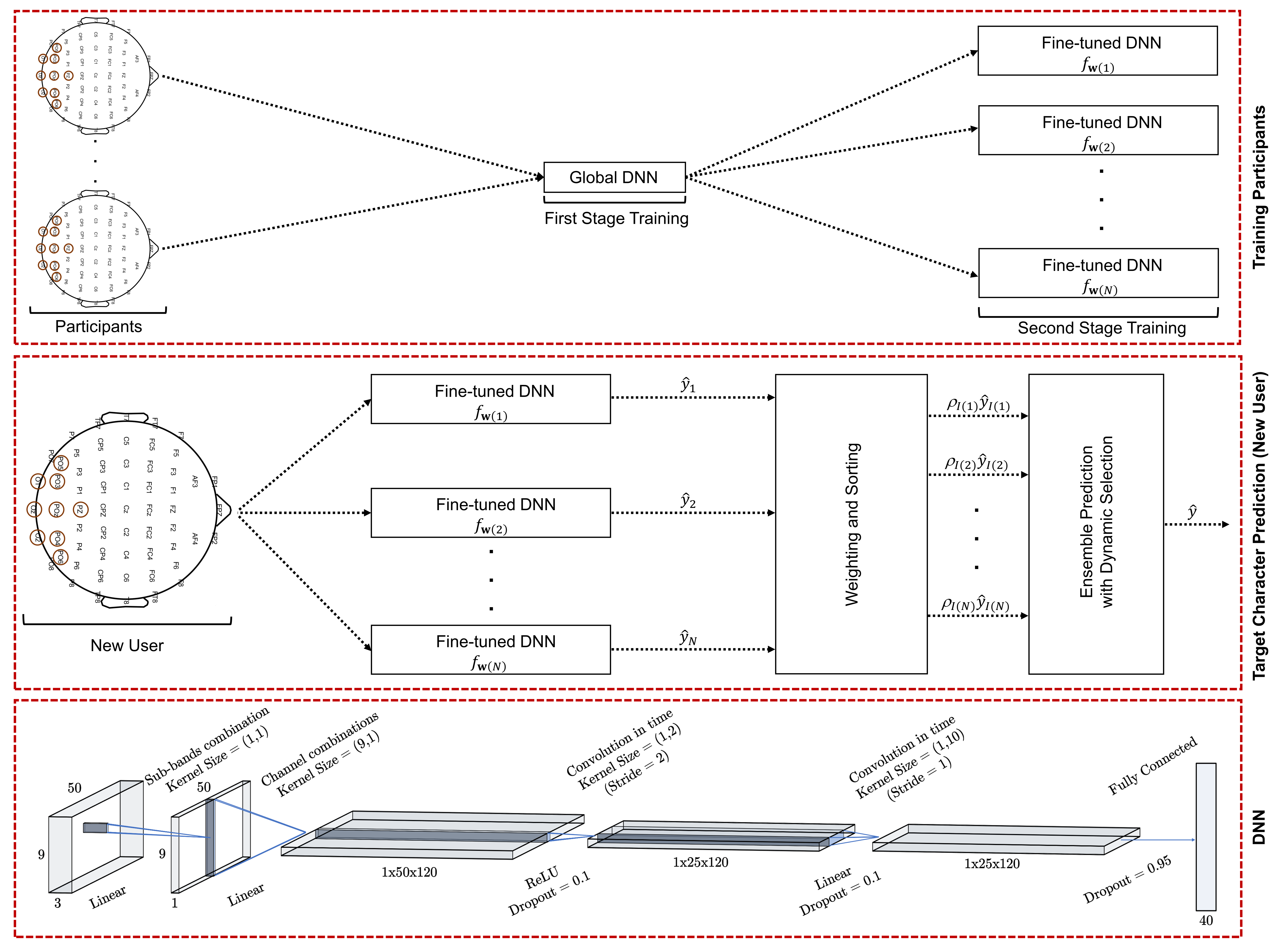}
\centering
\caption{Previously collected data of $N$ different participants are used to train a global DNN \cite{ourdnn}. This global DNN is fine-tuned to each participant which yields $N$ participant-specific fine-tuned DNNs. In the testing phase, the most representative $k$ fine-tuned DNNs are selected according to the statistical similarities between the new user and participants. Predictions of these $k$ DNNs are combined in a weighted manner to make the final target character prediction (target identification) for the new user.}
\label{fig:ensemble}
\end{figure*}

Our goal in this paper is to exploit this pre-existing data by using transfer and ensemble learning ideas so that a new user can start using the BCI system immediately (no user-specific training) with a high enough performance thanks to the transferred knowledge. For this purpose, we propose a highly novel target identification method for SSVEP BCI speller systems (cf. Fig. \ref{fig:ensemble}), which does not require user-specific training for a new user. Hence, as the main contribution of our study, we remove the hassle of long and tiring user-specific training periods present in the existing high performance spellers such as \cite{ourdnn} and \cite{trca}. Thus, the new user bears no hassle and her/his comfort is maximized. 

Our method (cf. Fig. \ref{fig:ensemble}) uses transfer and ensemble learning approaches in the sense that models that have been previously trained for different participants\footnote{In this paper, the word ``participants" refers to the participants of earlier EEG experiments conducted before a new user arrives, which provide some usable pre-existing SSVEP EEG speller data.} provide the ensemble and are transferred to the new user. {The target character is then identified based on a weighted combination of the predictions of  $k$ most representative models (i.e., the models of $k$ many participants whose data are statistically the most similar to that of the new user)  from the ensemble.} By using a dynamic selection, our method determines the best $k$ on the fly in the course of the target identification. Hence, we do not require any separate parameter tuning or parameter optimization or cross-validation steps. To construct the ensemble, we utilize a deep neural network (DNN) architecture that has been (Fig. \ref{fig:ensemble}) recently introduced in \cite{ourdnn}. On two publicly available and widely used datasets (the benchmark \cite{benchmark_Dataset} and BETA \cite{beta} datasets from $35$ and $70$ participants with $40$ target characters), we observe impressive $155.51$ bits/min (on the benchmark)  and $114.36$ bits/min (on BETA)  ITR performance figures. \textit{With these results, to the best of our knowledge, our proposed method is the best performing one among the most prominent literature methods that do not require user-specific training.}

The paper organization is as follows. We present the related work in Section \ref{sec:related_work}. Our problem description and method are in Section \ref{sec:goal} and Section \ref{sec:methodology}. We report our performance evaluations in Section \ref{sec:evaluations} and conclude in Section \ref{sec:CO}.

\section{Related Work} \label{sec:related_work}
The SSVEP BCI speller systems in the literature can be categorized into two groups. Unlike ours, the methods (e.g., \cite{ourdnn,osmanconf,trca}) in the first group require lengthy and tiring user-specific training.
Hence, this first group of methods are not appropriate for convenient daily life use, and are not comparable to our proposed method here. In our preliminary short conference proceeding (in Turkish) \cite{osmanconf}, we used the same ensemble of fined-tuned DNNs but combined all of them with weights given by a variant of the AdaBoost algorithm. In contrast, in this presented study, we combined only $k$ most representative fine-tuned DNNs based on a weighting devised through a novel correlation based similarity measure and dynamic selection, which is fundamentally different. Moreover, \cite{osmanconf} requires user-specific training thus belongs to the first group, whereas the presented study does not.

Like ours, the methods in the second group do not require any user-specific training. Either they are A) completely training-free,  or they rely on B) transferring knowledge from previous models/training participants as prior information. These methods allow the new user to comfortably start using the system immediately with no hassle. We explain this second group of methods below in two paragraphs.

In the PSDA method \cite{1499837}, the target character is decoded as the one corresponding to the frequency of the highest SNR signal from all possible stimulation frequencies. In the MEC method \cite{MEC}, the received EEG signals are linearly combined to cancel the noise components and increase the SNR. On the other hand, the CCA method \cite{CCA_method} determines the correlation between the SSVEP signals and  synthetically generated reference signals (composition of sinusoids of the stimulation frequencies) {by finding the channel and harmonic combinations that maximize the correlation}. The frequency that gives the maximum correlation is chosen as the target. Previous research has shown that the CCA method performs better than PSDA and MEC in terms of both the accuracy and ITR \cite{CCA_method,mec-cca-comp}. As an improved extension of CCA, a method based on a filter-bank (multiple band-pass filters) approach with canonical correlation analysis (FBCCA) is proposed in \cite{fbcca}. These band-pass filters are used to extract the sub-band components from the received EEG signals and then CCA is applied to each sub-band separately \cite{fbcca}. The target character is predicted by combining the results of CCA. 

In the tt-CCA method \cite{tt-CCA}, the target character is predicted based on the correlation between the received test signal and certain template signals. For each character, all of the existing participants' data (having the same label with the character in hand) are averaged to form a template signal. Then, CCA is applied between this template signal and the synthetic reference signal (composition of sinusoids of the corresponding frequency and its harmonics) so that the channel combination of the maximum correlation is selected. After completing this for all characters, the template/reference signals and channel combinations are transferred to the new user. While predicting the target character, tt-CCA combines three correlation coefficients for each character, and the one of the maximum combined correlation coefficient is the prediction. The first correlation coefficient is the maximum CCA coefficient between the new user instance and the reference signal. The channel combination of this maximum CCA coefficient combines the channels of the new user instance and the template signal, and then the correlation coefficient between them is calculated as the second. For the third coefficient, the same procedure is applied but with the transferred channel combination. Combined-tCCA \cite{comb-tCCA} follows a similar approach, except that it uses different correlation coefficients defined in \cite{extended-cca2}. Rather than employing user-specific coefficients, those of \cite{extended-cca2} are modified in \cite{comb-tCCA} by only using transferred signals and channel combinations. In the ttf-CCA method \cite{ttf-CCA}, the channel combinations are learned for each participant separately. After reducing the participant-specific channel combinations to certain common combinations, the correlation coefficients for each character between the new user instance and the corresponding template signal (similar to the templates of the tt-CCA method) are calculated. Maximum of these coefficients and another coefficient between the new user instance and the related synthetic template reveals the target prediction. 


\section{Problem Description}  \label{sec:goal}

We consider an $M$-character SSVEP BCI speller target identification ($M$-class classification) task for a multidimensional brain response\footnote{Vectors and matrices are shown with the bold mathematical font, whereas scalar variables are shown with the standard mathematical font.} (from a new user) $\matr{x}=[\matr{x}^{ (1)},\dots, \matr{x}^{(N_s)}] \in \mathbb{R}^{C \times N_t \times N_s}$, measured by EEG using $C$ channels/electrodes with $N_t$ sample points (i.e., $N_t = T \times F_s$ is the number of time samples with $F_s$ being the sampling frequency in Hz and $T$ being the signal length in seconds). Three band-pass filters are used ($N_s=3$) with no other preprocessing. Each of these filters has a different lower cut-off frequency $8\times s$ ($1\leq s \leq N_s$ is the filter index) and has the same upper cut-off frequency ($90$ Hz). We call this \textbf{SSVEP EEG signal $\matr{x}$} (to be target-identified) that is received from the new user during a spelling session as the \textbf{new user SSVEP EEG speller instance} or simply the \textbf{new user instance}. 

Our goal is to propose a highly novel SSVEP BCI speller target identification method (cf. Fig. \ref{fig:ensemble}) that is based on training an ensemble of DNNs with previously existing data from various participants. This ensemble is directly transferred to the new user with no user-specific training (i.e., with no lengthy and tiring preparation or calibration procedures or separate EEG experiments, all for algorithm training and system adaptation), and then a weighted combination of the target character predictions of a certain subset of the ensemble DNNs is used for making the final target identification, i.e., final target character prediction/classification for the new user. In addition to the target identification accuracy, we also use the (information transfer rate \cite{itr_cite}) $\text{ITR} = (\log_2M + P\log_2P + (1-P)\log_2 \left[ \frac{1-P}{M-1}\right])\frac{60}{T}$ to measure the performance of our proposed method, where $P$ is the target identification accuracy.



\textbf{Ensemble Constituent DNN:} We use the DNN architecture of \cite{ourdnn} (also shown in Fig. \ref{fig:ensemble}) to constitute our ensemble.
This DNN is an end-to-end system that consists of four convolutional layers and a single fully connected layer. The first convolutional layer is for sub-band combination while the second convolutional layer is for channel combinations. The third and fourth convolutional layers are used to extract features after applying filters. Finally, the fully connected layer predicts the target character by choosing the character of the highest probability returned by the softmax applied in the end. We use three sub-bands and nine channels (Pz, PO3, PO5, PO4, PO6, POz, O1, Oz, O2), whose physical locations on the skull are indicated in Fig. \ref{fig:ensemble}. 


\section{Method}  \label{sec:methodology}
\begin{enumerate}[label=\Alph*]
\item First, we construct an ensemble of target identification DNNs $\{f_{\matr{w}(n)}\}_{n=1}^N$ of size $N$, each of which is trained based on the labeled $M$-class SSVEP EEG speller data from a (previously existing) different participant $n$. Overall, we have $N$ participants, $1\leq n\leq N$. Here, $\matr{w}(n)$ parameterizes the $n$'th DNN with $f_{\matr{w}(n)}(\matr{x})=\hat{y}_n\in\{1,2,\cdots,M\}$ being the target character prediction for the SSVEP EEG signal $\matr{x}$ from the new user.    
\item Then, we identify the $k$ most representative DNNs $\{f_{\matr{w}(I(1))},f_{\matr{w}(I(2))},\cdots,f_{\matr{w}(I(k))}\}$ (here, $I(\cdot)$ is the indexing) after sorting the DNNs with respect to decreasing similarities between the new user's data and participants' data ($f_{\matr{w}(I(1))}$ is the most representative DNN or the data of the $I(1)$'th participant is statistically the most similar to that of the new user). We introduce a novel similarity measure $\rho_n$ for this purpose.
\item Finally, we obtain the final target identification for the SSVEP EEG signal $\matr{x}$ from the new user as a weighted linear combination of the $k$ most representative DNN predictions as $f(\matr{x})=\hat{y}=\argmax_{i \in \{1,\cdots,M\}} \sum_{j=1}^{k} \rho_{I(j)}\mathbbm{1}_{\{\hat{y}_{I(j)}=i\}}$. Here, $\mathbbm{1}_{\{\cdot\}}$ is the indicator function which outputs $1$ if the condition is satisfied (otherwise $0$) and $\rho_{I(j)}$'s are the combination weights that use our similarity measure. We also propose a novel dynamic selection algorithm for determining an appropriate value for the parameter $k$. Hence, as an additional merit that further improve the new user comfort, our technique does not require any cross-validation or parameter optimization for $k$ as we select it on the fly as a part of our final target character prediction.   
\end{enumerate}
\noindent
\textbf{Remark 1:} Our method does not require user-specific training for two reasons: i) Our ensemble of DNNs can be trained and constructed offline even before a new user arrives, and ii) all of our computations use only the new user instance (besides participants' data). Note that the new user instance is the SSVEP EEG signal that is received online during a spelling session and is the one to be target-identified. Hence, our method can be run online in real time, without requiring any offline data collections, and so any new-coming user can immediately start using our BCI system with no hassle.
\noindent
\textbf{Remark 2:} We point out that our method does successfully take into account two sources of statistical variations in the EEG signals: i) The person-to-person statistical variations by using only the $k$ most representative participant-specific DNNs (based on participant-to-new user statistical similarities) rather than using all of them, and ii) the within-person statistical variations by repeating the dynamic selection of the parameter $k$ for every new user instance.

Next, we expand on the details below in the same order.

\subsection{Training of the Ensemble $\{f_{\matr{w}(n)}\}_{n=1}^N$ of DNNs}

Our training strategy for obtaining the ensemble of DNNs is similar to the procedure in \cite{ourdnn} in one aspect since we also use two-stage training, but completely different in another aspect since we do not use user-specific training. i) In the first stage, a single global DNN is trained based on the entire set of labeled $M$-class SSVEP EEG speller data from pre-existing $N$ participants. ii) In the second stage, the global DNN is fine-tuned to each of the participants based only on her/his own data. Thus, an ensemble $\{f_{\matr{w}(n)}\}_{n=1}^N$ of fine-tuned DNNs is obtained. 
The first layer weights of the DNN architecture are initialized with unity, and the initialization of other layers' weights are sampled from Gaussian distribution with $0$ mean and standard deviation $0.01$. Also,  $0.1$, $0.1$ and $0.95$ dropout probabilities are applied between the second and third, third and fourth, and fourth and fifth layers, respectively. The network is trained by minimizing the categorical cross entropy loss $\frac{1}{D_b}\sum_{u=1}^{D_b} -\log(\matr{s}_{u,q_u}) + \lambda |\mathbf w|^{2}$, where $\lambda = 0.001$ is the weight of the L2 regularization, ${D_b}$ is the number of SSVEP EEG speller instances in the batch, $\matr{s}_{u,q_u}$ is the $q_u$'th index of the softmax output for the corresponding instance with the true label ${q_u}$, and $\matr{w}$ is the weights of the all layers in the DNN. This network setup and parameter initializations are same as the ones in \cite{ourdnn}. On the contrary, we emphasize that our target identification in this paper is an ensemble approach and is free of user-specific training/data as described, which is fundamentally different from \cite{ourdnn} where the target character prediction is made by a single network in a supervised manner that strictly requires labeled user-specific data.

\subsection{The Most Representative $k$ Participants: $\{f_{\matr{w}(I(j))}\}_{j=1}^k$} 

We use each participant-specific fine-tuned DNN in the ensemble $\{f_{\matr{w}(n)}\}_{n=1}^N$ to produce a target character prediction for an SSVEP EEG speller instance $\matr{x}$ of the new user, yielding an ensemble $\{f_{\matr{w}(n)}(\matr{x})=\hat{y}_n\}_{n=1}^N$ of predictions. 
However, since the EEG signal statistics are well know to significantly vary from one person to another, only a $k$-subset of our introduced ensemble is expected to be relevant to the final target identification. Namely, the DNNs of the participants that are more similar to the new user should provide more reliable predictions. Therefore, we devise a subset selection process for determining the most representative $k$ participants whose data are statistically the most similar (correlative) to that of the new user, which provide a $k$-subset of the DNN ensemble as $\{f_{\matr{w}(I(j))}\}_{j=1}^k$ with $I(\cdot)$ being the indexing (descending similarities). For this purpose, below we introduce a novel similarity measure between the participants and the new user.
 
Our similarity measure is based on the new user's SSVEP EEG speller instance, as well as the participants' instances that are correctly labeled by their respective fine-tuned DNNs. 
We start with defining an individual template $\overline{\matr{x}}_{n}^{i}$, which is the mean of the $n$'th participant's instances having the label $i \in \{1,2,\cdots,M\}$. Then, we also define a correlation-based metric, which is utilized in our similarity measure since that metric has been previously successfully employed in the literature for analysing the inter-participants statistical variations (cf. Section \ref{sec:related_work}). The utilized correlation metric is the summation of two correlation coefficients. The first one $\boldsymbol{\uprho}_{n,1}$ is the correlation between the $n$'th participant's template $\overline{\matr{x}}_{n}^{\hat{y}_n}$ and the new user instance $\matr{x}$. The second one $\boldsymbol{\uprho}_{n,2}$ is the correlation between the instance $\matr{x}$ and an artificial reference signal $\matr{Y}^{\hat{y}_n}\in \mathbb{R}^{2N_h \times N}$ that is formed for the ${\hat{y}_n}$'th character, where $N_{h}$ is the number of harmonics (we use $5$ harmonics: $N_{h}=5$). This artificial signal is generated for every character as the composition of sinusoids of the corresponding tagging frequency harmonics. The definition of this artificial signal can be found in Equation 1 of \cite{CCA_method}.
The first correlation coefficient $\boldsymbol{\uprho}_{n,1}$ can already be regarded as a direct similarity measure between the $n$'th participant and the new user; however, the second correlation coefficient $\boldsymbol{\uprho}_{n,2}$ (that we want to use) usefully takes into account the correctness of the $n$'th participant's DNN's prediction $\hat{y}_n$. 

It is nontrivial to calculate these correlations as the above-defined signals ($\matr{x}$, $\overline{\matr{x}}_{n}^{i}$, $\matr{Y}^{\hat{y}_n}$) are multi-dimensional, hence a reduction is necessary via combinations. Further, though such correlations can model the temporal statistics, the spatial statistics can also be probed using a channel combination ${\matr{w}_c^{(*)}} \in \mathbb{R}^{C \times 1}$ selected from the layers of the participants' DNNs (Fig. \ref{fig:ensemble}). Let the learned channel combination weights from the second layer of the $n$'th participant's DNN be denoted by ${\matr{w}_c}(n)$, i.e., ${\matr{w}_c}(n) = \{{\matr{w}_c^{(1)}}(n),{\matr{w}_c^{(2)}}(n),\cdots,{\matr{w}_c^{(N_{ch})}}(n) \} \in \mathbb{R}^{C \times N_{ch}}$, where $N_{ch}=120$ is the number of channel combinations (Fig. \ref{fig:ensemble}). These weights come after the sub-band combination weights ${\matr{w}_s}(n)$, i.e. ${\matr{w}_s}(n) = \{{\matr{w}_s^{(1)}}(n),{\matr{w}_s^{(2)}}(n),\cdots,{\matr{w}_s^{(N_{s})}}(n) \} \in \mathbb{R}^{N_{s} \times 1}$, in the employed DNN architecture. Therefore, we also combine the sub-bands of the new user instance $\matr{x}$ and the $n$'th participant's template $\overline{\matr{x}}_{n}^{\hat{y}_n}$ with ${\matr{w}_s}(n)$ in our similarity measure. This outputs $\underline{\matr{x}}_n = \sum_{s=1}^{N_s}{\matr{w}_s^{(s)}}(n)\matr{x}^{(s)}$ for the new user instance $\matr{x}$, and similarly $\underline{\overline{\matr{x}}}_{n}^{i}$ for the template. Likewise, a harmonic combination ${\matr{w}_Y} \in \mathbb{R}^{2N_h \times 1}$ is needed in the second correlation coefficient $\boldsymbol{\uprho}_{n,2}$ to combine the harmonics in $\matr{Y}^{\hat{y}_n}$. This harmonic combination  ${\matr{w}_Y}$ is found by CCA (canonical correlation analysis) for the given channel combination ${\matr{w}_c^{(*)}}$. 

Afterwards, we define the vector of aforementioned correlation coefficients
$\boldsymbol{\uprho}_n ({\matr{w}_c^{(*)}})$ as ($'$ is transpose below)
\[
\boldsymbol{\uprho}_n ({\matr{w}_c^{(*)}}) =
\begin{bmatrix}
    \boldsymbol{\uprho}_{n,1}({\matr{w}_c^{(*)}}) \\
    \boldsymbol{\uprho}_{n,2}({\matr{w}_c^{(*)}}) \\
\end{bmatrix} = 
\begin{bmatrix}
    \rho(({\matr{w}_c^{(*)}})^{'}\underline{\matr{x}}_n,({\matr{w}_c^{(*)}})^{'}\underline{\overline{\matr{x}}}_{{n}}^{\hat{y}_n})\\
    \rho(({\matr{w}_c^{(*)}})^{'}\underline{\matr{x}}_n,({\matr{w}_Y})^{'}Y^{\hat{y}_n}) \\
\end{bmatrix},
\]
\noindent
where the rightmost column keeps the regular correlation coefficients\footnote{The regular correlation coefficient between two $d$ dimensional signals/vectors $\matr{x}$ and $\matr{y}$ is: $\rho(\matr{x},\matr{y})=\frac{(\matr{x}-\sum_{u=1}^{d}\matr{x}_u)^T(\matr{y}-\sum_{u=1}^{N_s}\matr{y}_u)}{||\matr{x}-\sum_{u=1}^{d}\matr{x}_u||.||\matr{y}-\sum_{u=1}^{d}\matr{y}_u||}$.} 
between two one-dimensional signals, and the channel combination ${\matr{w}_c^{(*)}}$ is obtained from ${\matr{w}_c}(n)$ by maximizing the summation of the squared correlation coefficients:

\begin{equation}
    {\matr{w}_c^{(*)}} = \argmax_{\matr{w}_c^{(i)} \in \matr{w}_c(n)} \sum_{k=1}^{2}(\boldsymbol{\uprho}_{n,k}(\matr{w}_c^{(i)}))^2. \label{eq:best_channel_comb}
\end{equation}

\begin{algorithm}[t!]
\caption{Our Ensemble-DNN: Target identification of our method for a new user SSVEP EEG speller instance $\matr{x}$}\label{alg:f_extr}
\begin{algorithmic}[1]
\For {$n=1 \ to \ N$}
\State Get the prediction $\hat{y}_n=f_{\matr{w}(n)}(\matr{x})$ of the $n$'th DNN.
\State Get the template $\overline{\matr{x}}_{n}^{\hat{y}_n}$.
\State Get the sub-band combination $\underline{\matr{x}}_n = \sum_{s=1}^{N_s}{\matr{w}_s^{(s)}}(n)\matr{x}^{(s)}$ of the $n$'th DNN to $\matr{x}$.
\State Get the sub-band combination $\underline{\overline{\matr{x}}}_{n}^{\hat{y}_n} = \sum_{s=1}^{N_s}{\matr{w}_s^{(s)}}(n){\overline{\matr{x}}_{n}^{\hat{y}_n}}^{(s)}$ of the $n$'th DNN to $\overline{\matr{x}}_{n}^{\hat{y}_n}$.
\State Compute the similarity $\rho_n$ and sort.
\EndFor
\item Compute the $k$ value by using dynamic selection. 
\item Compute the final target prediction $\hat{y}$.
\end{algorithmic}
\end{algorithm}

Based on this setup, we define our similarity measure $\rho_n$ between the $n$'th participant and the new user as the sum of the squared correlation coefficients as
\begin{equation}
    \rho_n =  \sum_{k=1}^{2}(\boldsymbol{\uprho}_{n,k}(\matr{w}_c^{(*)}))^2. \label{eq:similarity}
\end{equation}
Hence, we obtain the set $\{f_{\matr{w}(I(j))}\}_{j=1}^k$ of most representative $k$ participants, provided that $I$ sorts the participants with respect to their similarities to the new user in the decreasing manner, i.e., $\rho_{I(j_1)}\geq \rho_{I(j_2)}$ if $j_1\geq j_2$ and $I(1)$ is the index of the most correlative/similar participant.

\subsection{Target Identification of the New User Instance $\matr{x}$}

In our proposed method, the following weighted linear combination of the predictions of the most representative $k$ participants is the final target identification of the new user SSVEP EEG speller instance $\matr{x}$:

\begin{equation}
    \hat{y}=\argmax_{i \in \{1,\cdots,M\}} \sum_{j=1}^{k} \rho_{I(j)}\mathbbm{1}_{\{\hat{y}_{I(j)}=i\}}. \label{eq:k_prediction}
\end{equation}
The parameter $k$ (number of most representative participants) can be chosen empirically or by cross-validation using participants' data. However, since the new user's statistics are different from the participants' statistics, using cross-validation might not be reliable. Choosing by hand empirically is not feasible as the best $k$ could vary for each new user and even for each new user instance. Instead, we propose a more sound strategy. After calculating the predictions for all $k\in\{1,2,\cdots, N\}$, we choose the $k$ of the prediction with the largest confidence that is defined as the weight difference between the most weighted character and the second most. The resulting $k$ can be different across not only new users but also new users' SSVEP EEG speller instances. We call this proposed $k$ selection strategy as \textbf{dynamic selection}. We emphasize that i) the proposed dynamic selection automatically determines the best $k$ for every new user instance on the fly while identifying the target character, and hence ii) our method removes the difficulty of finding the best $k$ by not requiring any separate parameter optimization/tuning step (no cross-validation or no manual tuning). This is another important contribution of our study. Finally, iii) by the introduced dynamic selection, we successfully address the person-to-person as well as the within-person statistical variations (from one new user instance to another) in the EEG signals. 

We summarize our method \textbf{Ensemble-DNN} step by step in Algorithm \ref{alg:f_extr}. Our \textbf{MATLAB code} is available for reproducibility at \textbf{https://github.com/osmanberke/Ensemble-of-DNNs}



\begin{figure}[t!]
    \centering
    \includegraphics[width=0.47\textwidth]{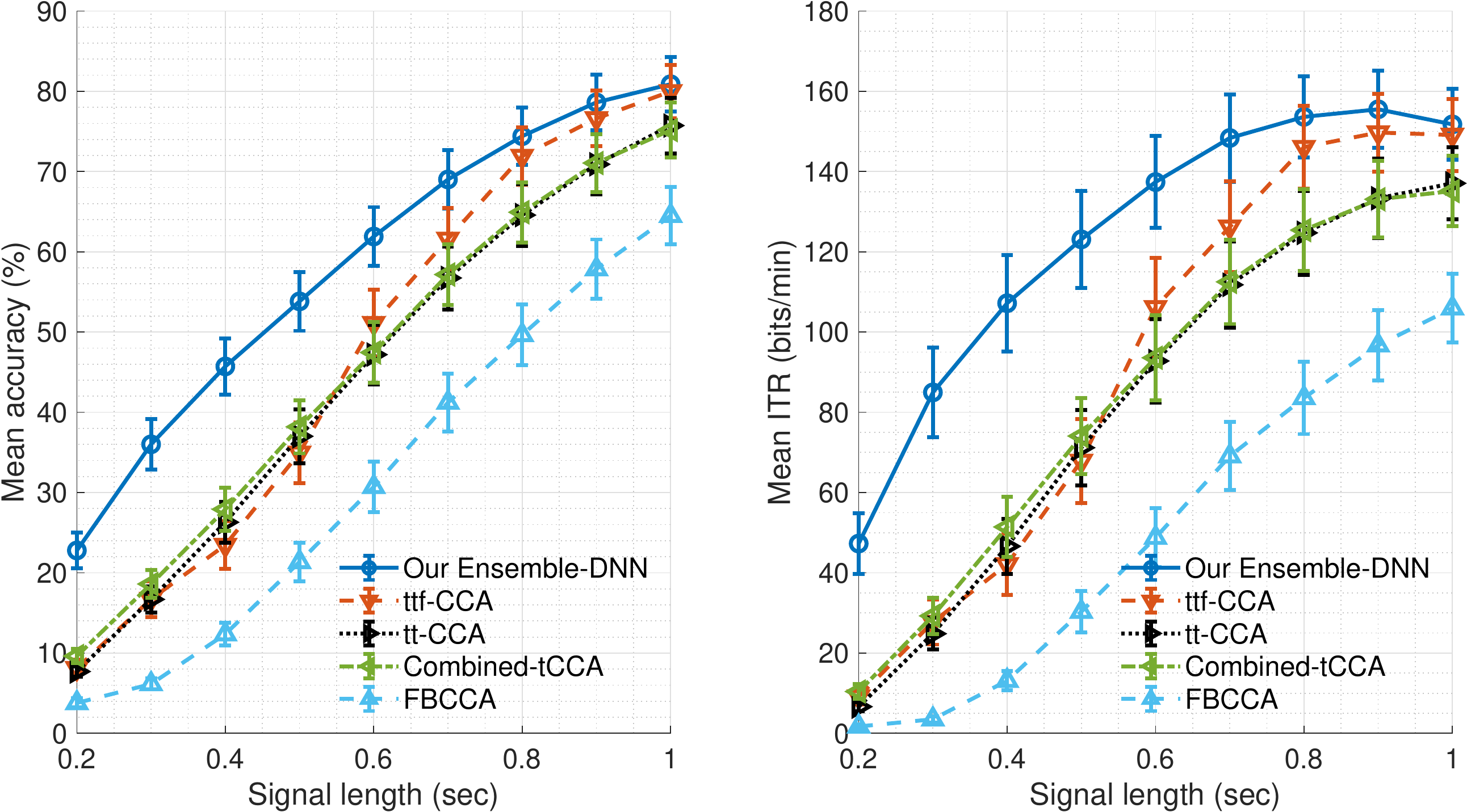}    \caption{The mean target identification accuracy on the left and the mean ITR on the right are presented across all $35$ participants in the benchmark dataset, together with the standard errors indicated by the bars.}
    \label{fig:bench_res}
    \centering
\end{figure}

\section{Performance Evaluations} \label{sec:evaluations}
 Our performance evaluations are based on the two publicly available and widely used benchmark \cite{benchmark_Dataset} and BETA \cite{beta} datasets which were generated with offline SSVEP BCI speller experiments. The benchmark experiments consist of $6$ blocks ($4$ blocks for BETA) and $35$ healthy participants ($70$ participants for BETA). Blocks include spelling of $M=40$ characters with one trial per each (for both the benchmark and BETA). In the benchmark experiments, the participant was shown a matrix ($5 \times 8$) of $40$ target characters each flickering with a unique frequency on a specific keyboard layout, whereas a QWERTY virtual keyboard was shown in the BETA experiments. Both datasets were collected by using EEG with $64$ channels. Since the BETA experiments were conducted outside of a laboratory environment, it is thus a more challenging dataset with lower SNR.

In our performance evaluations, for each dataset, we leave out one of the participants as the new user and use the rest ($N=34$ for benchmark and $N=69$ for BETA) as the pre-existing participants/data. This repeats in a leave-one-participant-out fashion for robust evaluations. Hence, there are $35$ repetitions for the benchmark dataset ($70$ for BETA), and each time a different participant becomes the new user. We compare our method against the tt-CCA, combined-tCCA, ttf-CCA, and FBCCA methods and report the mean target identification (i.e., multiclass classification) accuracy and ITR along with the standard errors (across leave-one-out repetitions). These methods are the most prominent and highest performing ones, among the comparable alternatives in the literature which do not require user-specific training like ours (cf. second group of methods in Section \ref{sec:related_work}). In each repetition, we only use the data of the new user to check the individual accuracy and ITR. Additionally, we compare our dynamic selection with the regular weighted combination (our Ensemble-DNN without dynamic selection) and the majority voting.

\subsection{Results and Discussion}
\begin{figure}[t!]
    \centering
    \includegraphics[width=0.47\textwidth]{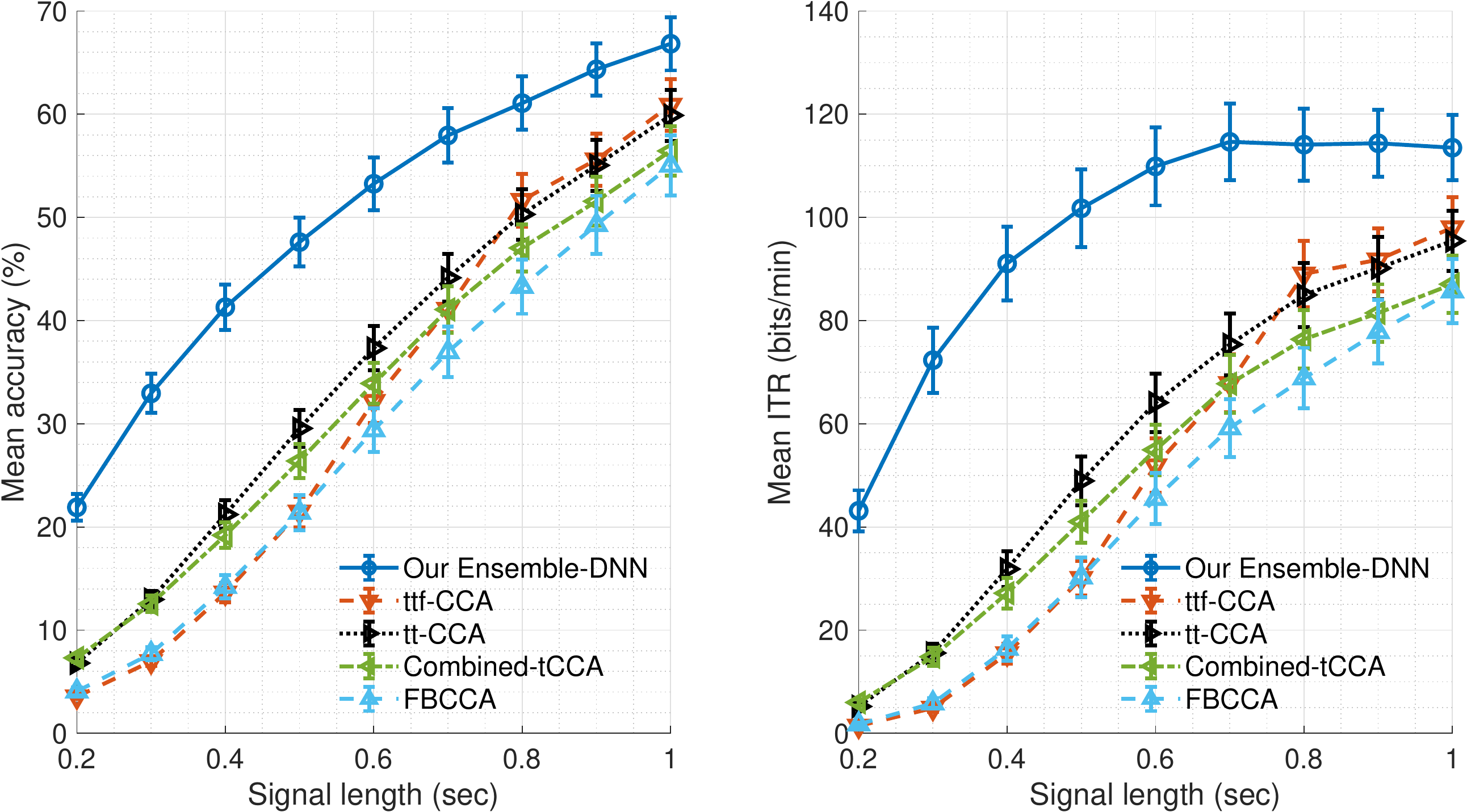}
    \caption{The mean target identification accuracy on the left and the mean ITR on the right are presented across all $70$ participants in the BETA dataset, together with the standard errors indicated by the bars.}
    \label{fig:beta_res}
    \centering
\end{figure}

On the benchmark dataset (Fig. \ref{fig:bench_res}), our proposed method (Ensemble-DNN) achieves $155.51$ bits/min mean ITR ($78.61\%$ mean target identification accuracy at $0.9$ seconds of signal length). We demonstrate in Fig. \ref{fig:bench_res} that our method significantly outperforms all the other compared methods within the range of stimulation of $[0.2,1]$ seconds. This is the most used range in the literature, probably because the prolonged exposure to flickering stimulus in the SSVEP experiments quickly becomes too tiring \cite{fatigue}. Hence, the shortest stimulation duration is certainly preferable where one also needs to achieve a sufficiently high ITR. In this sense, our proposed method is highly superior. 
On the BETA dataset (Fig. \ref{fig:beta_res}), we (Ensemble-DNN) achieve $114.64$ bits/min mean ITR ($57.95\%$ mean accuracy at $0.7$ seconds). Similar to the benchmark dataset, we demonstrate in Fig. \ref{fig:beta_res} that our method significantly outperforms the other methods on the BETA dataset as well. \textit{To the best of our knowledge, among all the methods in the literature that do not require any sort of user-specific training, data or adaptation from the new user, our ITR values (Ensemble-DNN) are the highest ever reported performance results on these datasets.}

\begin{figure}[t!]
    \centering
    \includegraphics[width=0.47\textwidth]{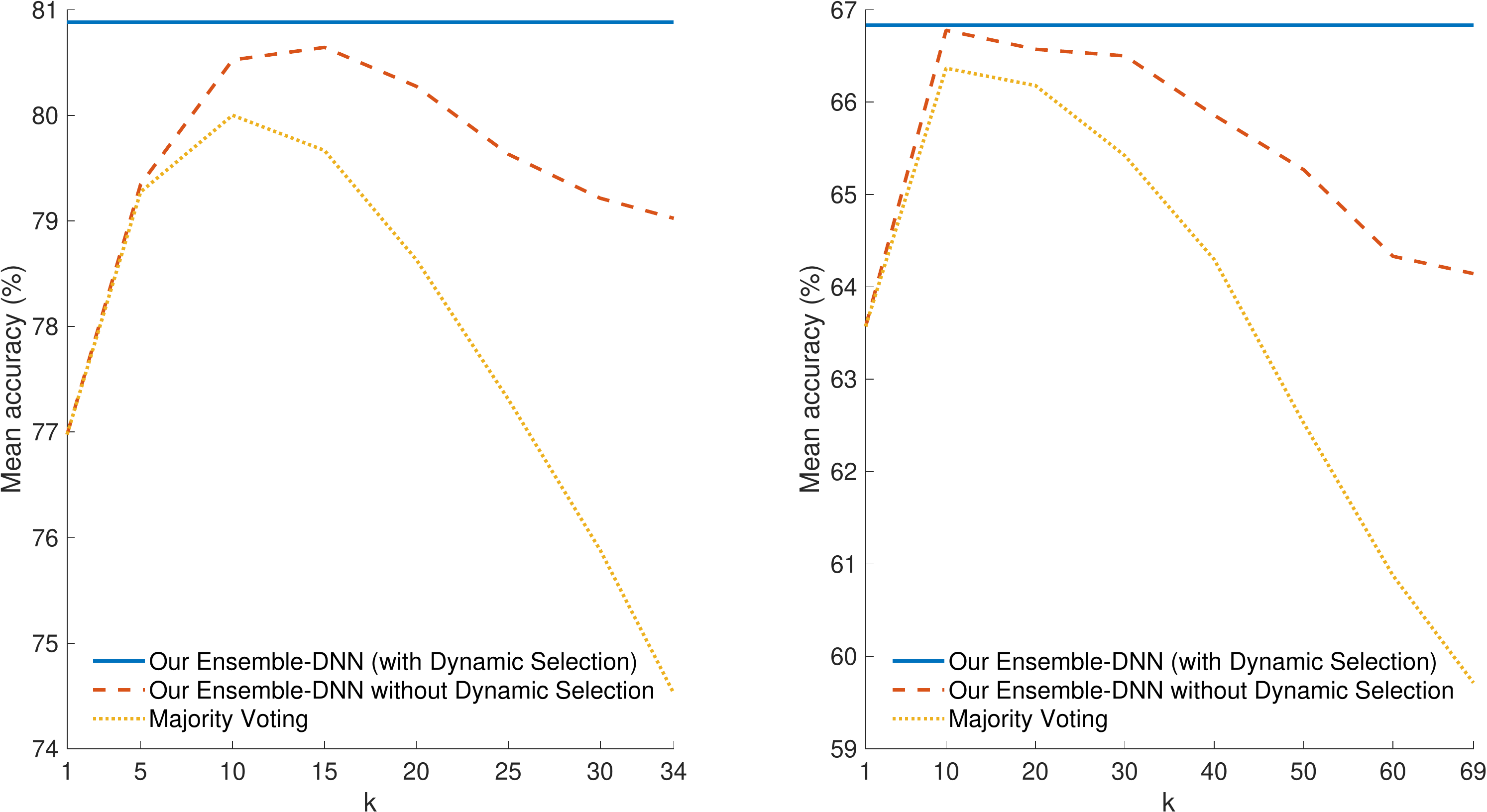}
        \caption{{The mean target identification accuracy of our Ensemble-DNN ({\bf\underline{with}} dynamic selection) is shown on the benchmark dataset (left) and the BETA dataset (right) at $1$ second signal duration, in comparison to the regular weighted combination (our Ensemble-DNN {\bf\underline{without}} dynamic selection) and majority voting for varying $k$.}}
    \label{fig:dynamic_selection}
    \centering
\end{figure}

Fig. \ref{fig:dynamic_selection} shows the mean target identification accuracies for varying $k$ choices (number of selected most representative participants) in our weighted combination in \eqref{eq:k_prediction}. This is our Ensemble-DNN without dynamic selection. Here, the same number $k$ of most representative participants is used for all the new users and for all their SSVEP EEG speller instances. We also compare in Fig. \ref{fig:dynamic_selection} with the majority voting as an alternative, where the majority of the target character predictions of all the participant-specific fine-tuned DNNs are used (to obtain a final prediction) without weighting. Finally, Fig. \ref{fig:dynamic_selection} also shows the mean target identification accuracy of our dynamic selection strategy for determining the $k$ value on the fly in the course of making a target character prediction. This is our Ensemble-DNN. Note that in our dynamic selection, the choice for $k$ is automatic and can be different across new users and across their SSVEP EEG speller instances. Hence, our method does also not require any manual parameter tuning or optimization or cross-validation. The results (Fig. \ref{fig:dynamic_selection}) reveal that our proposed dynamic selection strategy performs far better than the weighted combination of no selection and the majority voting, on both the benchmark and BETA datasets. 


\begin{figure}[t!]
    \centering
    \includegraphics[width=0.45\textwidth]{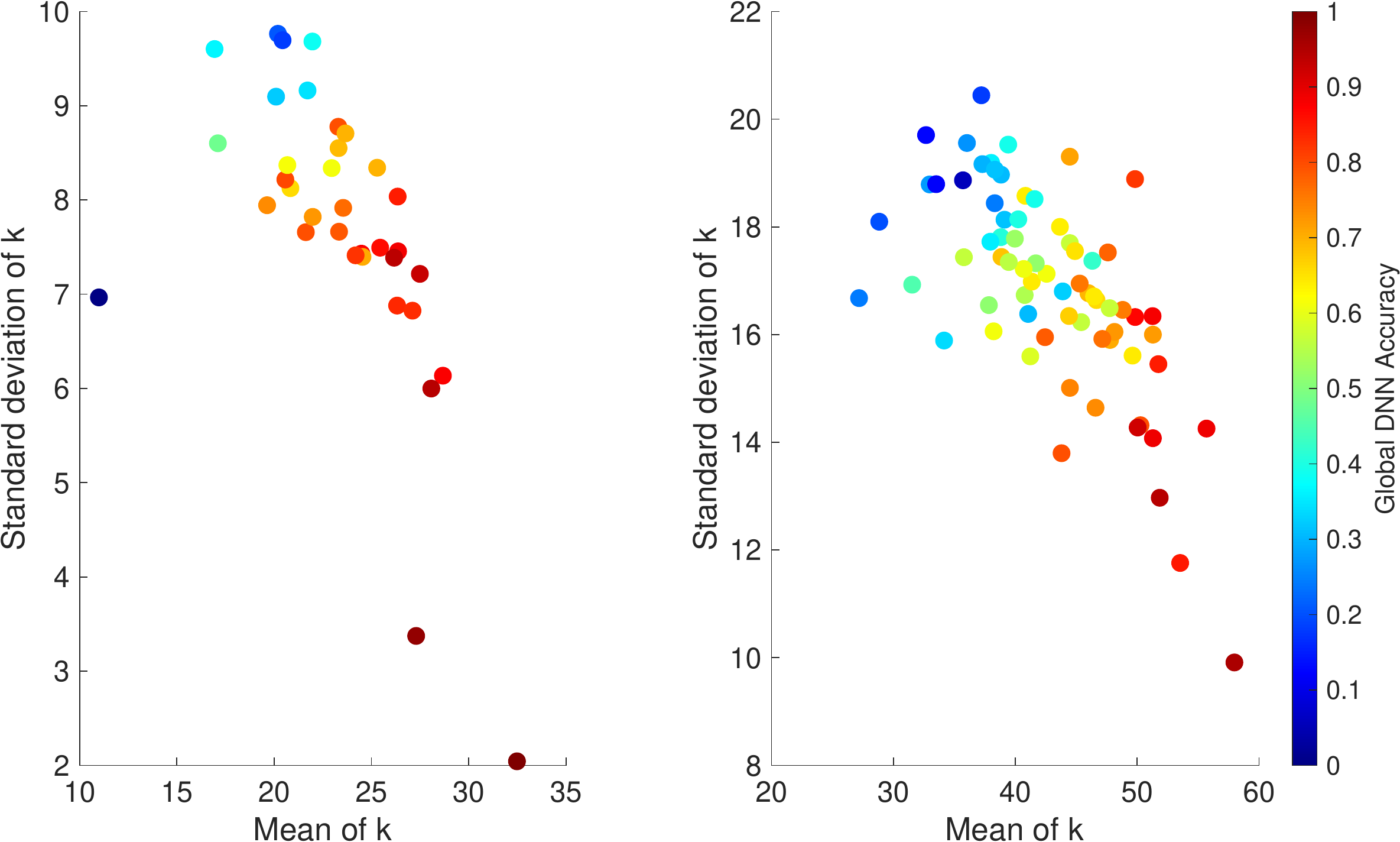}
    \caption{Each colored dot represents a benchmark (BETA) new user on the left (right), where the lighter (darker) color indicates higher (lower) global DNN target identification accuracy for that new user. The mean and standard deviation of $k$ (number of selected most representative participants) are across the SSVEP EEG speller instances of the corresponding new user and based on our proposed dynamic selection.}
    \label{fig:global_ensemble_acc}
    \centering
\end{figure}

Table \ref{table:EAA} provides comparisons with the no-ensemble approach of global DNN in which we directly use the target character predictions of global DNN without fine tuning (no second stage training) it to the participants. 
In Table \ref{table:EAA}, our proposed ensemble approach (Ensemble-DNN) greatly enhances the no-ensemble approach of global DNN in terms of the ITR and accuracy. This superiority is explained in Fig. \ref{fig:global_ensemble_acc} which presents two colored dot plots for the benchmark (left) and BETA (right) new users. 
We clearly observe that lighter colors accumulate on the bottom-right while darker colors accumulate on the upper left parts. This trend indicates the following. If the global DNN accuracy is high (low), then it means most participants are statistically similar (dissimilar) to the new user being tested, and in accordance, our dynamic selection method chooses more (less) participants, with high (low) mean and low (high) standard deviation for $k$, all expected. 
Consequently, our proposed Ensemble-DNN with dynamic selection relies only on the statistically similar participants and thus outperforms the no-ensemble approach of global DNN. 



\begin{table}[t!]
\caption{The mean target identification accuracy (on the first rows) and the mean ITR (on the second rows) comparisons between our proposed Ensemble-DNN and the global DNN (no-ensemble approach) are presented.}
 \small
\centering
\begin{tabular}{@ {\extracolsep{4pt}}ccccc}
\toprule   
{} & \multicolumn{2}{c}{Benchmark} &\multicolumn{2}{c}{BETA}  \\

\cmidrule{2-5} 
\centering 
 {} & Ensemble & Global & Ensemble & Global \\ 
 \midrule
{} & 80.88  &70.61          &66.83 & 56.58 \\
{1 sec} & 151.75 &123.31    &113.50& 88.12 \\
\midrule
{} & 78.61  &68.58          &64.34& 54.56 \\
{0.9 sec} & 155.51  &126.58 &114.36& 89.22 \\
\midrule
{} & 74.40  &64.42          &61.07& 51.94 \\
{0.8 sec} & 153.62 &123.54  &114.09& 89.21 \\
\bottomrule
\end{tabular}
\label{table:EAA}
\end{table}

\subsection{Statistical Significance Analyses}
For each $T\in \{0.2, 0.4, 0.6, 0.8, 1\}$, we conduct $4$ paired t-tests, pairing our proposed Ensemble-DNN method with the compared methods in Fig. \ref{fig:bench_res} and Fig. \ref{fig:beta_res}. We report unadjusted p-values, and the observed difference is called as ``statistically significant" (*) if the p-value is less than $\frac{0.05}{4}$ and ``statistically highly significant" (**) if the p-value is less than $\frac{0.05}{4\times 5}$. For the statistically significant case, we apply single Bonferroni correction by dividing $0.05$ by $4$, since for each $T$ there are $4$ comparisons. And for the statistically highly significant case, we apply double Bonferroni correction by $1/20$, since across all methods and $T$ choices there are $20$ comparisons. 

\textbf{In the case of the benchmark dataset:} In terms of the accuracy (Fig. \ref{fig:bench_res}), the least significant difference between our Ensemble-DNN method and the compared methods is observed with (1) ttf-CCA (**$p=9.40 \times 10^{-9}$) for $T=0.2$, (2) ttf-CCA (**$p=7.37 \times 10^{-9}$) for $T=0.4$, (3) ttf-CCA (**$p=5.82 \times 10^{-6}$) for $T=0.6$, (4) ttf-CCA ($p=0.27 \times 10^{-1}$) for $T=0.8$, (5) ttf-CCA ($p=0.48$) for $T=1.0$. For $T=0.8$, the difference with ttf-CCA is not significant; but it is highly significant (**) with all the others. For $T=1.0$, the difference with ttf-CCA is not significant; but it is significant (*) with tt-CCA and highly significant (**) with all the others. In terms of ITR (Fig. \ref{fig:bench_res}), the least significant difference between our Ensemble-DNN method and the compared methods is observed with (1) ttf-CCA (**$p=1.91 \times 10^{-6}$) for $T=0.2$, (2) ttf-CCA (**$p=1.38 \times 10^{-7}$) for $T=0.4$, (3) ttf-CCA (**$p=5.72 \times 10^{-6}$) for $T=0.6$, (4) ttf-CCA ($p=0.18 \times 10^{-1}$) for $T=0.8$, and (5) ttf-CCA ($p=0.43$) for $T=1$. For $T=0.8$, the difference with ttf-CCA is not significant; but it is highly significant (**) with all the others. For $T=1.0$, the difference with ttf-CCA is not significant; but it is significant (*) with tt-CCA and highly significant (**) with all the others.

\textbf{In the case of the BETA dataset:} In terms of the accuracy (Fig. \ref{fig:beta_res}), the least significant difference between our Ensemble-DNN method and the compared methods is observed with (1) FBCCA (**$p=6.89 \times 10^{-21}$) for $T=0.2$, (2) FBCCA (**$p=3.21 \times 10^{-22}$) for $T=0.4$, (3) FBCCA (**$p=4.96 \times 10^{-17}$) for $T=0.6$, (4) FBCCA (**$p=2.03 \times 10^{-11}$) for $T=0.8$, (5) tt-CCA (**$p=1.43 \times 10^{-5}$) for $T=1.0$. In terms of ITR (Fig. \ref{fig:beta_res}), the least significant difference between our Ensemble-DNN method and the compared methods is observed with (1) Combined-tCCA (**$p=6.55 \times 10^{-16}$) for $T=0.2$, (2) Combined-tCCA (**$p=1.05 \times 10^{-18}$) for $T=0.4$, (3) tt-CCA (**$p=1.21 \times 10^{-16}$) for $T=0.6$, (4) ttf-CCA (**$p=2.21 \times 10^{-11}$) for $T=0.8$, and (5) tt-CCA (**$p=2.48 \times 10^{-6}$) for $T=1$. In terms of both the accuracy and ITR, the difference with all of the compared methods for all $T$'s is always statistically highly significant (**). 

\section{Conclusion} \label{sec:CO}


For SSVEP BCI speller systems to become practical and commonplace, we shall not require any additional data from a new-coming user for algorithm training and system adaptation. However, most existing target identification techniques require that burden of lengthy and tiring EEG data collection processes to achieve high ITR performances. Our proposed method in this work ensures practicality by using the already-existing literature datasets from various previous EEG experiment participants. We first train participant-specific target identifier DNNs based on participants' own pre-existing data, and then combine the $k$ most representative (in the sense of statistical similarities between the data of participants and the data of new user) DNNs to predict the target character for a given new user instance during a spelling session. Thus, our method does not require any additional training data from a new user who wants to use the system immediately with no hassle. The proposed method achieves significant improvements in the ITR performances compared to the state-of-the-art alternatives while promoting the wide-spread BCI use in daily lives.

\section{Acknowledgement}
This work was supported by The Scientific and Technological Research Council of Turkey under Contract 121E452. We thank Can Aksoy, Emirhan Koc, Suayb Arslan and Yigit Catak for their participation in our discussions and their help in our MATLAB implementations and performance evaluations.

\bibliographystyle{ieeetr}
\bibliography{IEEEabrv,references}

\end{document}